\documentclass[runningheads]{llncs}
\usepackage{graphicx}
\usepackage{url}
\usepackage{color}
\usepackage{xcolor}
\usepackage{soul}
\usepackage{graphicx}
\usepackage[utf8]{inputenc} % allow utf-8 input
\usepackage[T1]{fontenc}    % use 8-bit T1 fonts
\usepackage{lmodern}
\usepackage{hyperref}       % hyperlinks
\usepackage{url}            % simple URL typesetting
\usepackage{booktabs}       % professional-quality tables
\usepackage{amsfonts}       % blackboard math symbols
\usepackage{nicefrac}       % compact symbols for 1/2, etc.
\usepackage{microtype}      % microtypography
\usepackage{wrapfig}
\usepackage{graphicx}
\usepackage{amsmath}
\usepackage[ruled,vlined]{algorithm2e}
\usepackage{cite}

\newcommand{\EQ}{\begin{equation}}
\newcommand{\NQ}{\end{equation}}
\newcommand{\ER}{\begin{eqnarray}}
\newcommand{\NR}{\end{eqnarray}}
\newcommand{\ERS}{\begin{eqnarray*}}
\newcommand{\NRS}{\end{eqnarray*}}
\newcommand{\bit}{\begin{itemize}}
\newcommand{\ben}{\begin{enumerate}}
\newcommand{\eben}{\end{enumerate}}
\newcommand{\ebit}{\end{itemize}}

\newcommand{\bc}{{\bf c}}

\newcommand{\bu}{{\bf u}}

\newcommand{\bx}{{\bf x}}

\newcommand{\bC}{{\bf C}}

\newcommand{\bS}{{\bf S}}

\begin{document}
\title{Few-shot Image Recognition with Manifolds}
\author{Debasmit Das \and
J. H. Moon \and
C. S. George Lee}
\authorrunning{D. Das et al.}
\institute{School of Electrical and Computer Engineering, Purdue University, \\West Lafayette, IN, USA\\
\email{\{das35,moon92,csglee\}@purdue.edu}\\}
\maketitle              % typeset the header of the contribution
\begin{abstract}
In this paper, we extend the traditional few-shot
learning (FSL) problem to the situation when the source-domain
data is not accessible but only high-level information in the
form of class prototypes is available. This limited information
setup for the FSL problem deserves much attention due to its
implication of privacy-preserving inaccessibility to the source-domain
data but it has rarely been addressed before. Because of
limited training data, we propose a non-parametric approach to
this FSL problem by assuming that all the class prototypes are
structurally arranged on a manifold. Accordingly, we estimate the
novel-class prototype locations by projecting the few-shot samples
onto the average of the subspaces on which the surrounding
classes lie. During classification, we again exploit the structural
arrangement of the categories by inducing a Markov chain on
the graph constructed with the class prototypes. This manifold
distance obtained using the Markov chain is expected to produce
better results compared to a traditional nearest-neighbor-based
Euclidean distance.
To evaluate our proposed framework, we
have tested it on two image datasets – the large-scale ImageNet
and the small-scale but fine-grained CUB-200. We have also studied
parameter sensitivity to better understand our framework.\\
\end{abstract}
\section{Introduction}

Deep learning has produced breakthrough in many areas 
like computer vision~\cite{krizhevsky2012imagenet,he2016deep}, 
speech recognition~\cite{amodei2016deep}, 
natural language processing~\cite{choemnlp} etc., 
mainly due to the availability of lots of labeled data, 
complex neural network architectures and %advanced but 
efficient training procedures. 
Even though these deep learning models are trained on large labeled datasets, 
they still fail to generalize to new classes or environments. 
Humans, on the other hand, can quickly recognize new objects from very few samples. 
They do that by using their previously obtained \emph{knowledge} and apply it to new situations. 
This difference in the way machines and humans learn provides motivation 
to carry out research on few-shot learning (FSL). Accordingly, most few-shot learning methods strive  for a transfer-learning approach
where they extract useful transferable knowledge from data-abundant base classes 
and use it to recognize data-starved novel classes.

\begin{wrapfigure}{r}{0.3\textwidth}
  \begin{center}
    \includegraphics[width=0.3\textwidth]{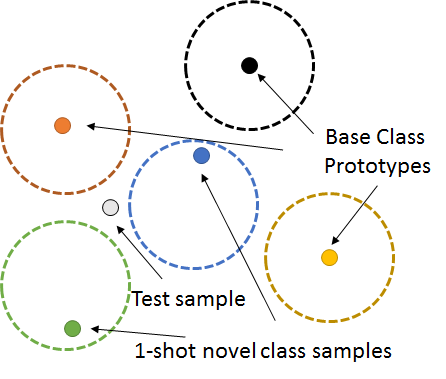}
  \end{center}
  %\vspace{-8pt}
  \caption{In this FSL setting, the base-class prototypes are known but not the novel-class prototypes. The spread of the classes (dashed boundaries) are also unknown.}
\label{realfsl}
%\vspace{-10pt}
\end{wrapfigure}

Most previous methods in FSL assumed that abundant labeled data is available 
across the base categories from which a robust and generalizable knowledge representation can be learned. However, in certain situations, it is difficult to have access to all the labeled data 
of these base categories due to privacy restrictions and/or 
inefficiency in maintaining such a large database. 
%It might be because there are privacy restrictions on the data or 
%that it might be inefficient to maintain such a large historical database. 
Hence, alternatively class exemplars or prototypes can be retained. These prototypes summarize the class information by averaging over the data samples without revealing sensitive information about the data. For example, the prototype of the dog category can be the arithmetic mean of all the dog sample features. Previous work on \emph{hypothesis transfer learning} (HTL)~\cite{jie2011multikern,tommasi2014PAMI,kuzborskij2017scalable} assumed access to base class models, where recognition performance would depend on the choice of these models. As a result, this HTL-based setting does not allow for fair comparison. On the other hand, our proposed setting involving base-class prototypes allow for fair comparison where performance depends only on the data and the proposed transfer learning approach.
This proposed FSL setting is depicted in Fig.~\ref{realfsl}.

% In this setting, it allows having access to only a single test sample at a time rather than all the unlabeled test data. However, this assumption has been relaxed in recent semi-supervised-learning-based attempts to address FSL~\cite{ren2018meta,garcia2017few}.

Previous FSL methods cannot be applied to this proposed restrictive setting. This is because they assume that lots of labeled data are available from the source domain. Consequently, they use neural-network-based parametric approaches.  The neural-network-based approaches can be categorized depending on the type of transferable knowledge 
extracted from the base categories and encoded in the neural network architecture: (a)\emph{Metric-learning} methods~\cite{snell2017prototypical,sung2018learning,vinyals2016matching} 
learn a metric space; (b)\emph{Meta-learning} methods~\cite{finn2017model,ravi2016optimization,andrychowicz2016learning} learn the learning procedure; (c)\emph{Generative} methods~\cite{WangCVPR2018a,hariharan2017low,mehrotra2017generative} generate data for the novel classes. However, the neural-network-based parametric models might severely overfit if we only have limited data in the form of class prototypes available from the source domain. Hence, in order to solve this restrictive FSL problem, it is natural to seek a non-parametric approach.

In this paper, we address this restricted FSL setting by formulating it 
as a case of \emph{ill-sampling}. 
As depicted in Fig.~\ref{realfsl}, 
the correct locations of the base category prototypes are known but that of the novel categories are unknown. This is because the few-shot data from a novel category 
might be sampled from the periphery of the class distribution. 
These non-representative samples when used for classification will result in poor recognition performance. Therefore, a non-parametric-based  prior is used to produce a biased estimate of the novel-class prototype location. 

\begin{figure}[htb]
%\vspace*{-0.1in}
  \begin{center}
    \includegraphics[width=\textwidth]{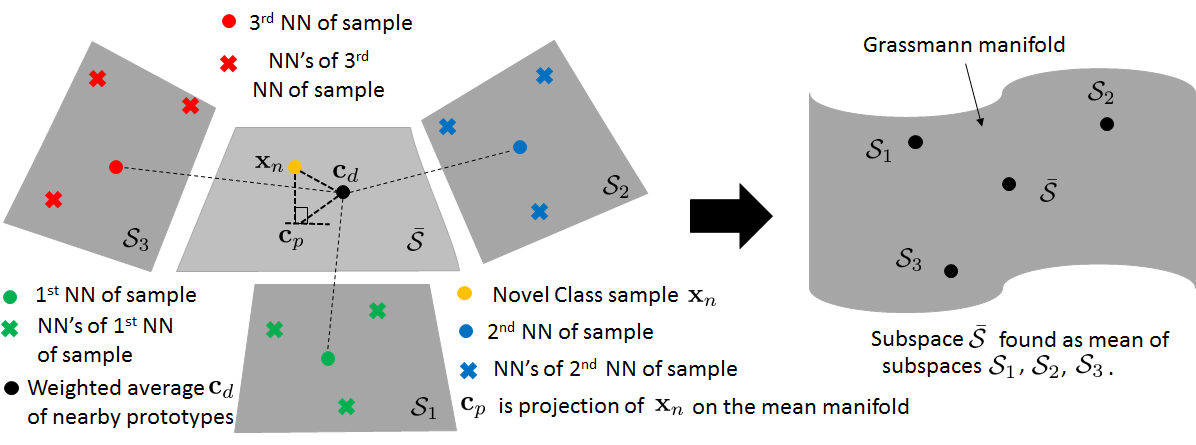}
  \end{center}
  \caption{The surrounding subspaces $\mathcal{S}_1$, $\mathcal{S}_2$ and $\mathcal{S}_3$ for a novel-class sample $\bx_n$ are found by using its Nearest Neighbors (NN) and the NNs of its NNs. Their mean can be calculated to obtain the subspace $\bar{\mathcal{S}}$ on which the novel-class sample $\bx_n$ is projected to obtain $\bc_p$. The weighted average $\bc_d$ of the nearby prototypes are also used to obtain the novel-class prototype.}
  \label{fig:subspace}  
%\vspace{-15pt}
\end{figure}

For the non-parametric-based prior, we find inspiration from the idea~\cite{basri2003lambertian} that data samples from one class lie on a low-dimensional subspace. Therefore, we can consider all the classes as a collection of piece-wise linear subspaces. This set of subspaces can be considered as an approximation of a non-linear manifold close to which the class-prototypes lie. This manifold serves as a prior to estimate the location of the novel-class prototype. The subspace near the novel-class prototype is found by calculating the mean of the subspaces on which the nearby base classes lie. The subspace on which the nearby base classes lie is again found 
using their nearest neighbors as shown in Fig.~\ref{fig:subspace}. 
Finally, the novel-class sample can be projected onto 
the mean subspace to obtain the novel-class prototype.

Once the novel-class prototypes are estimated, 
one can use the nearest-neighbor (NN) approach to assign a test sample to a class 
based on the Euclidean distance to all the prototypes. 
However, the estimation procedure for the novel-class prototype might still be error prone. Hence, there is a need to exploit the structural arrangement of the manifold containing all the classes to assign a class to a test sample. The neighboring class locations can provide better estimate of class-prototype distances. This can be achieved by constructing a graph using all the class prototypes and then using equilibrium probability of an induced absorbing Markov-chain process to output the most probable class. Finally, to validate the proposed approach, we perform experiments and analyses on this framework to set a benchmark for future research.

\vspace{-12mm}
% Our major contributions in this paper are as follows: 
% (a) We introduce the restrictive setting of few-shot learning where only base category prototypes are available; 
% (b) We propose the idea of classes lying on a nonlinear manifold 
% and use piece-wise linear subspaces to estimate novel-class prototype locations; 
% (c) The structural arrangement of classes on a manifold is used to predict the category 
% of a test sample by inducing a Markov-chain process on the corresponding graph; 
% (d) We consequently perform experiments and analyses so as to set a benchmark for future research.
%\vspace{-10pt}
\section{Proposed Approach}
In this section, we describe the proposed framework, which consists of two steps - estimating novel-class prototypes using the manifold approach and classification using the Markov chain method.
%\begin{figure}
%  \begin{center}
%   \includegraphics[width=\textwidth]{subspace.png}
% \end{center}
%  \caption{The surrounding subspaces $\mathcal{S}_1$, $\mathcal{S}_2$ and $\mathcal{S}_3$ for a novel-class sample $\bx_n$ are found by using its Nearest Neighbors (NN) and the NNs of its NNs. These subspaces when orthonormalized are represented as points on the Grassmann manifold. Their mean can be calculated to obtain the subspace $\bar{\mathcal{S}}$ on which the novel-class sample $\bx_n$ is projected to obtain $\bc_p$. The weighted average $\bc_d$ of the nearby prototypes are also used to obtain the novel-class prototype.}
%  \label{fig:subspace}
%  \vspace{-15pt}
%\end{figure}
\subsection{Estimating Novel-Class Prototypes}
Consider that we have access to the base-category prototypes 
collected in the form of a matrix $\mathbf{C} \in \mathbb{R}^{n_b \times d}$, 
where $n_b$ is the number of base prototypes and $d$ is the dimensionality of the feature space. 
Let the one-shot sample from a novel class be $\mathbf{x}_n \in \mathbb{R}^d$. 
Our goal is to estimate the prototype location $\mathbf{c}_n \in \mathbb{R}^{d}$ of the novel class. 
Our assumption is that all the base and novel-class prototypes, i.e., all rows of $\mathbf{C}$ and $\mathbf{c}_n$ lie close to a non-linear manifold. 
Since the mathematical expression of the manifold is unknown, 
we express it as a collection of piece-wise linear subspaces. 
First, we find the surrounding classes of the novel class by finding $r$ nearest-neighboring prototypes 
of the novel-class sample $\mathbf{x}_n$. 
Let these nearest-neighboring prototypes be denoted as $\mathbf{c}_{ni} \in \mathbb{R}^d$ 
for $i \in \{1,2, ... ,r\}$. 
For each of the $r$ neighboring prototypes, 
we find $q$ neighboring prototypes. 
These new prototypes can be expressed as $\mathbf{c}_{nij} \in \mathbb{R}^d$ 
for $j \in \{1,2, ... ,q\}$ and $i \in \{1,2, ... ,r\}$. 
Hence, $\mathbf{c}_{nij}$ represents the $j^{th}$ nearest neighbor 
of the $i^{th}$ nearest neighbor of the novel-class sample $\mathbf{x}_n$. 
To represent the non-linear manifold, we form $r$ linear subspaces 
using the $r$ nearest neighbors of the novel sample 
as well as the $q$ nearest neighbors of each of the $r$ prototypes. 
The linear subspace $\mathcal{S}_i$ corresponding to the $i^{th}$ nearest neighbor 
of $\bx_n$ is represented as a column space {such that} 
$\mathcal{S}_i \equiv [\mathbf{c}_{ni}~\vdots~\mathbf{c}_{ni1}~\vdots~\mathbf{c}_{ni2}~\vdots~....~\vdots~\mathbf{c}_{niq}]$. 
The linear subspace can be orthonormalized to obtain $\mathcal{S}^{\perp}_i$ 
and the operation can be repeated for all the $r$ nearest neighbors. 
The net result is $r$ linear subspaces with dimensionality $(q+1)$ 
surrounding the novel-class sample $\mathbf{x}_n$. 
In the example in Fig.~\ref{fig:subspace}, we chose $r=3$ and $q=3$. 
These linear subspaces represent linearized localized versions of the non-linear manifold 
on which the class prototypes lie. 
The subspace on which the novel-class prototype lies close to can be found 
by averaging these surrounding $r$ subspaces $\mathcal{S}^{\perp}_i$ for $i \in \{1,2,...,r\}$. 
For finding the average of these orthonormal subspaces, 
we use the concept of Grassmann manifold.

A Grassmann manifold $\mathcal{G}(n,l)$ for $n,l > 0$ is the topological space composed of all $l$-dimensional linear subspaces embedded in an $n$-dimensional Euclidean space. A point on the Grassmann manifold is represented as an $n \times l$ orthonormal matrix $\mathbf{S}$ whose columns span the corresponding linear subspace $\mathcal{S}$. It is represented as: $\mathcal{G}(n,l)=\{\text{span}(\mathbf{S})$: $\mathbf{S} \in \mathbb{R}^{n \times l}, \mathbf{S}^{T}\mathbf{S}=\mathbf{I}_{l}\}$, where $\mathbf{I}_{l}$ is a $l\times l$-dimensional identity matrix and superscript $T$ indicates matrix transpose.

Following the definition, the $r$ orthonormal subspaces $\mathcal{S}^{\perp}_i$ for $i \in \{1,2,...,r\}$ are points lying on a $\mathcal{G}(d,q+1)$ Grassmann manifold. 
The average of these points on the Grassmann manifold will represent the linear subspace 
to which the novel-class prototype lies close to. 
The average of these points is found using the \emph{extrinsic mean}. 
For a set of points on the Grassmann manifold $\mathcal{G}(d,q+1)$, 
the extrinsic mean is the point that minimizes the Frobenius-norm-squared difference 
of the projections of the points onto the space of $(q+1)$ ranked $d \times d$ matrices. 
Therefore, the optimization problem for finding the extrinsic mean $\bar{\mathbf{S}}$ is 
\begin{equation}
\label{eq:karcher}
\begin{gathered}
\underset{\mathbf{S}^{{T}}\mathbf{S}=\mathbf{I}}{\text{argmin}}\sum_{i=1}^{r}d(\mathbf{S}_i,\mathbf{S})^2, \quad \text{where}  \quad d(\mathbf{S}_i,\mathbf{S})=\frac{||\bS\bS^T-\bS_i\bS^T_i||_\mathcal{F}}{\sqrt{2}}.
\end{gathered}
\end{equation}
Here $||\cdot||_{\mathcal{F}}$ is the Frobenius norm.
Let ${\mathbf{S}}^*$ be the solution to the optimization problem~\eqref{eq:karcher}, which can be found using eigenvalue decomposition.
${\mathbf{S}}^*$ is the spanning matrix of the extrinsic mean of the surrounding subspaces $\mathcal{S}^{\perp}_i$'s. Setting the extrinsic mean $\bar{\mathbf{S}}={\mathbf{S}}^*$, we project the novel-class sample $\mathbf{x}_n$ onto the subspace spanned by the matrix $\bar{\mathbf{S}}$. The projected point $\mathbf{c}_p$ can be obtained as $\mathbf{c}_p=\bar{\mathbf{S}}\bar{\mathbf{S}}^{+}\mathbf{x}_n$, where $\bar{\mathbf{S}}^{+}=(\bar{\mathbf{S}}^{{T}}\bar{\mathbf{S}})^{-1}\bar{\mathbf{S}}^{{T}}$. {Superscripts $-1$ and $+$ indicate matrix inverse and matrix pseudo-inverse, respectively.}

We also consider the direct contribution of the surrounding class prototypes 
into estimating the novel-class prototype. 
If $\mathbf{C}_r \in \mathbb{R}^{r \times d}$ consists of the $r$ nearest neighbors 
of the novel-class sample $\mathbf{x}_n$, 
then their contribution $\mathbf{c}_d$ to the novel-class prototype location can be found 
using the equation $\mathbf{c}_d=\mathbf{C}^{{T}}_r\mathbf{p}_d$,
where $\mathbf{p}_d \in \mathbb{R}^{r} $ is the probability vector formed by carrying out the exponential mapping of the Euclidean distances of the class prototypes to $\mathbf{x}_n$, followed by normalization. Hence, the contributions $\mathbf{x}_n$, $\mathbf{c}_p$ and $\mathbf{c}_d$ can be used to estimate the novel-class prototype location $\mathbf{c}_n$ as
\begin{equation}
\label{eq:weigh}
\mathbf{c}_n=\alpha_2[\alpha_1 \mathbf{x}_n+ (1-\alpha_1)\mathbf{c}_p]+(1-\alpha_2)\mathbf{c}_d,    
\end{equation}
where $\alpha_1$, $\alpha_2 \in [0,1]$ are scalar weights that are manually set. In case $\mathbf{x}_n$ is very close to the novel-class prototype, $\alpha_1=\alpha_2 \approx 1$ will produce optimal classification performance.
 \vspace{-8mm} 
%\vspace{-10pt}
\begin{wrapfigure}[18]{r}{0.4\textwidth}
 \begin{center}
    \includegraphics[width=0.4\textwidth]{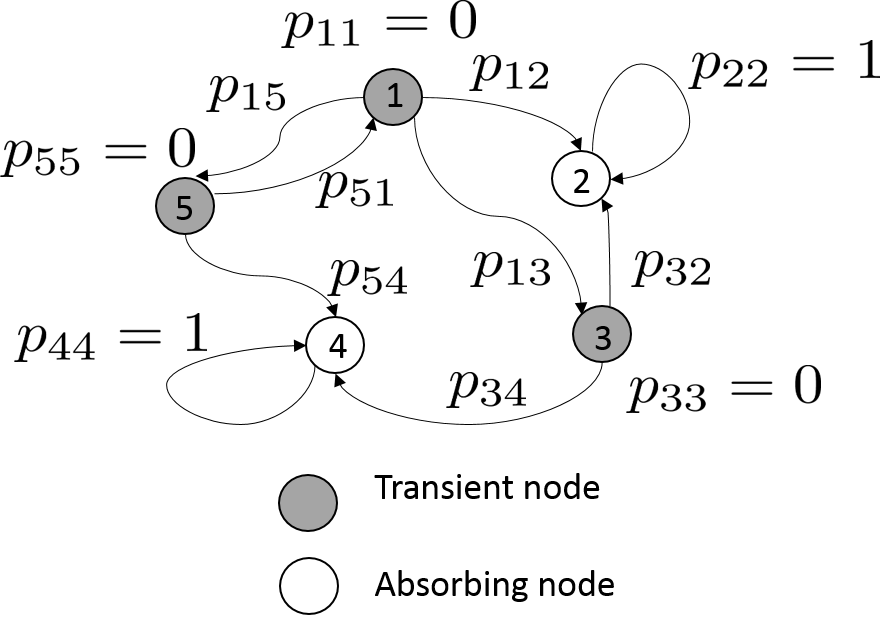}
    %\vspace{-10pt}
\end{center}
  \caption{Possible transitions are shown with directed arrows. The transition probability from {a state} $i$ to a state $j$ is $p_{ij}$. The transient state and the absorbing state have self-transition probabilities $p_{ii}$ as $0$ and $1$, respectively.}
\label{markov}
\end{wrapfigure}
\subsection{Classification using Absorbing Markov Chain}

Once the class prototype locations of the novel classes are known, 
the structural arrangement of the prototypes of both the base and novel classes 
are again used to recognize a test sample to obtain a more
informed decision about the classification. The structural arrangement of the classes is represented using a $k'$-nearest-neighbor ($k'$-NN) graph, where each node represents a class prototype.
The $k'$-NN graph formulation allows nodes to only be connected to its $k'$-NN nodes. 
The weights between the nodes are defined using the exponential of the negative Euclidean distances. 
Upon defining this graph, an absorbing Markov-chain process is induced on it. 
Each state of the Markov chain corresponds to a node in the graph and therefore a category. 
The transition probability from {a state} $i$ to a state $j$ is found as $p_{ij}=w_{ij}/\sum_{l}w_{il}$, where $w_{il}$ is the weight connecting {nodes} $i$ and $l$. In the absorbing Markov chain process, there are two kinds of states - \emph{transient} and \emph{absorbing}. The transient state and the absorbing state have self-transition probabilities $p_{ii}$ as $0$ and $1$, respectively. This suggests that a random walker on a graph cannot stay on the transient node for the next step but for the absorbing node it will stay there forever. An example of an absorbing Markov chain is given in Fig.~\ref{markov}, 
where the arrows represent the possible transitions from one node to another. 
Overall, the Markov chain is represented using the transition matrix $\mathbf{P}$, 
which models the dynamics of the process. 
Using $\mathbf{P}$, the {Markov-chain equations} are described as follows:
\begin{equation}
\label{eq:markov}
\begin{gathered}
 \mathbf{u}^{t+1}=\mathbf{u}^{t}\mathbf{P}, \quad \text{where} \quad \mathbf{P}=\begin{bmatrix} \mathbf{T}_{n_t \times n_t} & \mathbf{A}_{n_t \times n_a} \\ \mathbf{0}_{n_a \times n_t} & \mathbf{I}_{n_a \times n_a} \end{bmatrix}, 
\mathbf{P}^{\infty}=\begin{bmatrix} \mathbf{0}_{n_t \times n_t} & (\mathbf{I}-\mathbf{T})^{-1}\mathbf{A} \\ \mathbf{0}_{n_a \times n_t} & \mathbf{I}_{n_a \times n_a} \end{bmatrix}.
\end{gathered}
\end{equation}
$\mathbf{u}^{t}$ and $\mathbf{u}^{t+1}$ are the states of the process at instants $t$ and $t+1$, respectively, and they are represented as a probability vector over all the states.
$\mathbf{T}$ describes the transition probabilities from one transient state to another.
$\mathbf{A}$ describes the transition probabilities from transient states to absorbing states. $n_t$ and $n_a$ are the number of transient and absorbing states, respectively, and 
the zero and identity matrices $\mathbf{0}_{n_a \times n_t}$ and $\mathbf{I}_{n_a \times n_a}$ 
imply that the process cannot leave the absorbing state. 
Our goal is to find the equilibrium state $\mathbf{u}^t$ as $t \rightarrow \infty$ 
for a given initial state $\mathbf{u}^0$. 
Accordingly, $\mathbf{u}^{\infty}=\mathbf{u}^0\mathbf{P}^m$ as $m \rightarrow \infty$. 
The closed-form solution of $\mathbf{P}^{m}$ as $m \rightarrow \infty$ is treated as $\mathbf{P}^{\infty}$. Using this formulation, the equilibrium state probabilities can only be distributed among the absorbing states with zero probabilities on the transient states.

The initial state $\mathbf{u}^0$ is calculated using the Euclidean distances of the test sample 
to all the base- and novel-class prototypes and normalizing it to obtain a probability vector. 
Using the absorbing Markov-chain formulation in Eq.~\eqref{eq:markov}, we choose the novel categories and base categories 
as transient and absorbing states, respectively, 
to obtain the most probable base category from $\mathbf{u}^{\infty}$. Then, we interchange the order of absorbing and transient states to obtain the most probable novel category. 
In the final step, we apply {one nearest neighbor} approach on the test sample 
to choose the most probable class among the most probable base and novel categories 
obtained in the previous steps. 
The overall procedure from the novel-class prototype estimation 
to the Markov-chain-based prediction for a test sample 
is given in \emph{Algorithm 1}. 
In case we have multiple samples for a novel class, $\bx_n$ is set as the mean of these samples.
\begin{algorithm}[h]
\SetAlgoLined
% \KwResult{Write here the result }
%  initialization\;
%\vspace*{-0.1in}
 \textbf{Given:} Base category prototypes $\bC \in \mathbb{R}^{n_b \times d}$, Novel class one-shot samples $\bx_n, n \in \{1,2,...,n_{nov}\}$, Test sample $\bx_{te}$.\\ 
 \textbf{Parameters:} $r, q, k', \alpha_1, \alpha_2$\\
 \textbf{Goal:} Classify $\bx_{\text{te}}$ into one of the $n_b+n_{nov}$ categories\\
 \textbf{Step 1} \emph{Estimate class prototype for each novel class}\\
\textbf{for} each novel class $n \in \{1,2,...,n_{nov}\}$\\
\quad Obtain $r$ nearest base prototypes for $\bx_n$ to form $\bC_r \in \mathbb{R}^{r \times d}$\\
\quad Obtain $q$ nearest base prototypes for each of the $r$ base prototypes\\
\quad Obtain orthonormal subspaces $\mathcal{S}^{\perp}_i$ for $i \in \{1,2,...,r\}$ using
the $q$ neighbors \\
\quad $\bar{\mathcal{S}} \leftarrow \texttt{ExtrinsicManifoldMean}(\mathcal{S}^{\perp}_1,\mathcal{S}^{\perp}_2,...,\mathcal{S}^{\perp}_r)$\\
\quad Project $\bx_n$ onto $\bar{\mathcal{S}}$ to obtain $\bc_p$ followed by $\bc_d$ \\
\quad Obtain novel-class prototypes as $\mathbf{c}_n \leftarrow \alpha_2(\alpha_1 \mathbf{x}_n+ (1-\alpha_1)\mathbf{c}_p)+(1-\alpha_2)\mathbf{c}_d$ \\
\textbf{end for}\\
\textbf{Step 2} \emph{Predict class of test sample} $\bx_{te}$ \\
Construct $k'$-nearest-neighbor graph with $n_b$ base prototypes and $n_{nov}$ novel prototypes as nodes.\\
Find initial probability vector $\bu_0$ using distance of $\bx_{te}$ to all the prototypes.\\
Construct Markov chain and obtain most probable base class.\\
Construct Markov chain and obtain most probable novel class.\\
Use nearest neighbor to obtain the most probable class among the two.
\caption{Proposed two-step FSL procedure using manifolds.}
\end{algorithm}
\section{Experiments and Discussions}
\subsection{Dataset Description}
%\vspace{-10pt}
To evaluate our proposed approach, we used two image recognition datasets -- ImageNet and CUB-200. 
Originally, the ImageNet dataset consists of 21K categories of which we used 1000 for our experiments. These 1000 categories are accordingly split into base and novel classes. The CUB-200 is a fine-grained dataset that consists of 200 categories of different bird species. 
Of these 200 classes, we used a total of 150 of which 100 are base and 50 are novel classes. 
For both datasets, the image features used were the 2048-dimensional ResNet-101~\cite{he2016deep}.
%\vspace{-0.1in}
\subsection{Effects of varying the number of classes and samples}
%\vspace{-10pt}
In this section, we study how recognition performance is affected by 
the number of categories and the number of samples per category in the base and novel datasets. 
For training purposes, we used the prototypes of the base categories and the few-shot samples 
from the novel categories. For evaluation purposes, test samples from both base and novel categories were used. Recognition performance is reported as class-wise averaged accuracy. 
This ensures that major classes do not dominate the performance and minor classes 
containing less number of samples are not ignored. 
It is noted that performing cross-validation is impossible since we do not have access to enough data to be held {out} as a validation set 
and therefore results were reported by fixing the hyper-parameters. 
For evaluation, we used the following models: (NA) The No-adaptation baseline which consists of just using nearest neighbor on the few-shot sample mean; (M1) It uses nearest neighbor on the estimated novel-class prototypes; (M2) It uses the Markov-chain-based manifold distance on the few-shot sample mean; (Oracle) It assumes access to novel-class prototypes and uses nearest neighbor for prediction; (M1+M2) involves computing the manifold distance on the estimated class prototypes. 

For the first set of experiments, we used the ImageNet dataset with 800 base and 200 novel categories and studied the effect of changing the number of shots per novel category. The results were taken over 10 trials and reported in Table~\ref{tab:imgsample}. 
From the results, it is seen that M1 improves the recognition performance over the no-adaptation baseline but the difference diminishes as the number of shots increases. This is because for the novel categories, the few-shot mean becomes closer to the prototype location as the number of shots increases. Also, the contribution of M2 over the baseline or over M1 is incremental. 
This can be attributed to the fact that the ResNet-101 features are not trained 
using the manifold-based distance and there is a mis-match 
between the training and testing evaluation measures. 
The standard error reduces with the increasing number of shots 
because of reduced variance in the few-shot mean and eventually the estimated prototype over the trials. 
We repeated the same experiment for the CUB-200 dataset, the results of which are reported in Table~\ref{tab:cubsample}. 
In this case, we have 100 base and 50 novel categories, all of which are fine-grained. 
As a result, the recognition performance is poorer compared to ImageNet, even though CUB-200 has lesser number of categories. Still, the observed recognition performance has a pattern similar to that of the ImageNet dataset. However, there is no reduction in the standard error with increasing shots. This can be attributed to larger overlap between the fine-grained classes of CUB-200.
\begin{table}[h]
\begin{center}
    \caption{Average accuracy results over 10 trials of the ImageNet dataset with 800 base and 200 novel categories as the number of shots per novel category is changed. Standard error is shown in the parentheses. The hyper-parameter setting is $r=20,q=20,k'=3,\alpha_1=0.9,\alpha_2=0.7$.}
    \label{tab:imgsample}
\begin{tabular}{cccccc}
\cline{1-6} \hline
 & 1 shot       & 2 shot        & 5 shot       & 10 shot        & 20 shot            \\ \cline{1-6} \hline
\textbf{NA}                                                     & 64.31 (0.05) & 67.60 (0.05)  & 71.09 (0.03) & 72.24 (0.02) & 72.89 (0.01)  \\ 
\textbf{M1}                                                & 66.58 (0.05) & 69.67 (0.05)  & 71.62 (0.03) & 72.31 (0.02) & 72.91 (0.01)  \\ 
\textbf{M1}+\textbf{M2}                                              & 66.91 (0.05) & 69.88 (0.05)  & 72.05 (0.03) & 72.72 (0.02) & 72.97 (0.02) \\ 
\textbf{M2}                                                & 65.21 (0.05)  & 67.98 (0.05) & 71.33 (0.02) & 72.07 (0.02)  & 72.60 (0.01)  \\ 
\textbf{Oracle}                                                 & 73.3         & 73.3          & 73.3         &  73.3              & 73.3               \\\cline{1-6}  \hline
\end{tabular}
\end{center}
 \vspace{-8mm} 
\end{table}
\begin{table}[h]
\begin{center}
    \caption{Accuracy results over 10 trials of the CUB-200 dataset with 100 base and 50 novel categories as the number of shots per novel category is changed. The hyper-parameter setting is $r=20,q=20,k'=5,\alpha_1=0.5,\alpha_2=0.5$.}
    \label{tab:cubsample}
\begin{tabular}{cccccc}
\cline{1-6} \hline
          & 1 shot       & 2 shot            & 5 shot            & 10 shot           & 20 shot           \\ \cline{1-6} \hline
\textbf{NA}       & 43.40 (0.12) & 45.28 (0.13)  & 51.19 (0.18)  & 55.70 (0.20) & 58.16 (0.16) \\
\textbf{M1}    & 45.91 (0.23) & 48.80 (0.20) & 51.90 (0.16) & 55.94 (0.18) & 57.63 (0.14) \\ 
\textbf{M1}+\textbf{M2}  & 46.13 (0.28) & 49.01 (0.22) & 52.13 (0.11)  & 55.57 (0.17) & 58.67 (0.12) \\
\textbf{M2}    & 43.81 (0.11) & 45.86 (0.15) & 51.45 (0.15) & 55.91 (0.18)  & 58.31 (0.14) \\
\textbf{Oracle}    & 60.51     & 60.51           & 60.51           & 60.51           & 60.51\\ \cline{1-6}\hline    
\end{tabular}
\end{center}
 %\vspace{-8mm} 
\end{table}
For the next set of experiments, we considered the performance change on the ImageNet dataset 
for the 1-shot setting as the numbers of base and novel categories are varied. 
We considered two such scenarios. In the first case, the total number of categories was fixed at 1000 
while the proportion of base categories was changed. 
This setting considers less number of base categories compared to novel categories 
and it has rarely been studied in previous work. 
The results of this setting are reported in Table~\ref{tab:imgratio}. 
From the table, it can be seen that M1 improves over NA by a large margin (9 points) 
especially when the number of base categories is very less (ratio of 0.1). 
This is alluded to our assumption that all the class prototypes have a structural arrangement on a manifold. Therefore, the use of this structure is especially beneficial in the few-class regime. 
However, the difference between M1 and NA decreases mainly due to  more base categories  and lesser difficult novel categories. 
After that, we considered the experimental setting where the total number of categories was varied 
but the proportion of base and novel categories was kept the same at 4:1, 
the results of which are reported in Table~\ref{tab:imgtotal}. 
The results show that the upper bound of the recognition performance, i.e., the oracle performance, decreases with an increase in the number of categories. 
This is because classification becomes more difficult as the number of categories increases. 
As expected, M1 performs better compared to NA and the contribution of M2 is incremental.

Our proposed few-shot learning setting is new and therefore we do not have previous work 
to compare and benchmark against. 
However, we can study whether our approach can improve existing relevant work. Prototypical  networks~\cite{snell2017prototypical} consider the mean of the few-shot samples to represent class prototypes. Therefore, there is a possibility of obtaining the class prototypes using our manifold-based approach and further improving the performance. Accordingly, we tested the contribution of M1 and M2 over prototypical networks on \emph{miniImageNet}, which is a subset of the ImageNet dataset. The results are shown in Table~\ref{tab:protonet}. In the table, $K$-way $N$-shot implies that $K$ novel categories are sampled per testing episode with $N$ samples per category. From the results, it is clear that M1 improves the performance, however M2 declines it. 
This is mainly because of the discrepancy between the Euclidean distance metric 
used during training ProtoNets and the manifold-based distance metric used during testing.
\begin{table}[h]
\begin{center}
    \caption{Accuracy results over 10 trials of the ImageNet dataset for the 1-shot setting as the ratio of number of base categories to the total number of categories is changed. ($x$-b, $y$-n) implies $x$ base and $y$ novel categories.}
    \label{tab:imgratio}
\begin{tabular}{ccccc}
\cline{1-5} \hline
          & 0.1 (100-b, 900-n)& 0.2 (200-b,800-n)   & 0.4 (400-b, 600-n) & 0.6 (600-b, 400-n)            \\ \cline{1-5} \hline
\textbf{NA}  & 38.29 (0.30)      & 39.59 (0.29)   & 46.50 (0.18)  & 55.28 (0.11)     \\ 
\textbf{M1}  & 47.70 (0.28) & 49.65 (0.30)    & 54.65 (0.23)  & 60.71 (0.12)         \\ 
\textbf{M1}+\textbf{M2} & 47.89 (0.28)  & 50.33 (0.29)   & 55.13 (0.23)  & 61.34 (0.11)     \\ 
\textbf{M2}  & 38.36 (0.30)  & 39.68 (0.29)   & 46.57 (0.19)   & 55.31 (0.11)       \\ 
\textbf{Oracle} & 73.3    & 73.3 & 73.3 & 73.3   \\ \hline
 \cline{1-5}   
\end{tabular}
\end{center}
\end{table}
%\vspace*{0.1in}h
\begin{table}[h]
\begin{center}
    \caption{Accuracy results over 10 trials of the ImageNet dataset for the 1-shot setting as the total number of categories is changed but keeping the ratio of base categories to novel categories as 4:1.}
    \label{tab:imgtotal}
\begin{tabular}{ccccc}
\cline{1-5} \hline
          & 50 (40-b, 10-n) & 100 (80-b, 20-n) & 200 (160-b, 40-n) & 500 (400-b, 100-n) \\ \cline{1-5} \hline
\textbf{NA}        & 81.76 (0.54)    & 75.98 (0.34)     & 70.25 (0.16)      & 67.43 (0.10)  \\
\textbf{M1}   & 86.10 (0.45)     & 79.92 (0.36)     & 73.50 (0.16)      & 69.59 (0.08)  \\
\textbf{M1}+\textbf{M2} & 86.43 (0.47)   & 80.61 (0.51)     & 74.09 (0.34)      & 70.41 (0.08)  \\
\textbf{M2}  & 82.52 (0.53)    & 75.48 (0.39)     & 70.88 (0.23)      & 67.44 (0.10)  \\
\textbf{Oracle}    & 92.20            & 86.76       & 80.85          & 76.87            \\ \hline
 \cline{1-5}   
\end{tabular}
\end{center}
 \end{table}
%\vspace*{0.1in}
\begin{table}[h]
\begin{center}
    \caption{Few-shot classification accuracies on the miniImageNet dataset averaged over 600 test episodes for different ways and shots. 95\% confidence intervals are shown in the parentheses.}
    \label{tab:protonet}
\begin{tabular}{ccccc}
\cline{1-5} \hline
          & 5-way 1-shot & 5-way 5-shot & 20-way 1-shot & 20-way 5-shot  \\ \cline{1-5} \hline
\textbf{ProtoNet}        & 47.21 (0.69)    & 63.62 (0.61)    & 20.51 (0.46)   & 35.20 (0.59)  \\
\textbf{ProtoNet+M1}   & 48.79 (0.51)     & 65.67 (0.56)     & 21.93 (0.62)      & 35.66 (0.53)  \\
\textbf{ProtoNet+M2} & 41.36 (0.47)   & 57.48 (0.43)     & 16.94 (0.57)  & 31.52 (0.64)   \\\hline
 \cline{1-5}   
\end{tabular}
\end{center}
%\vspace{-20pt}
% \vspace{-8mm} 
\end{table}
%\vspace{-10pt}
\subsection{Parameter Sensitivity Studies}
\begin{figure}[h]
%  \begin{center}
\centering
    \includegraphics[width=0.8\textwidth]{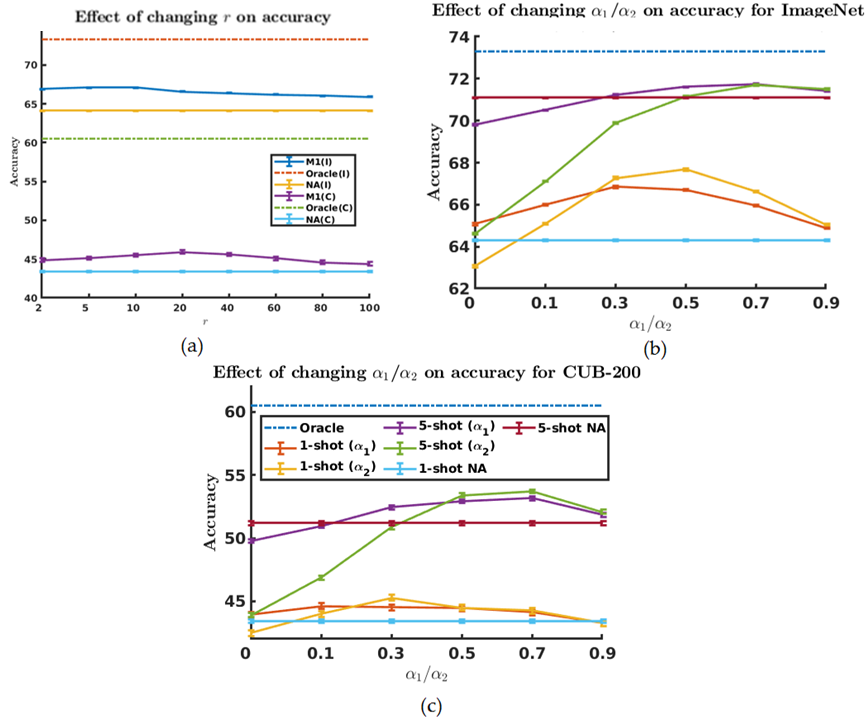}
%  \end{center}
  %\vspace{-10pt}
  \caption{(a) Effect of the number of subspaces $r$ on recognition performance for both ImageNet (I) and CUB-200 (C). Effect of $\alpha_1$ and $\alpha_2$ on 1-shot and 5-shot recognition performance for  (b) ImageNet and (c) CUB-200. Legends of (c) hold for (b) as well. $\alpha_1$ in parenthesis suggests that $\alpha_1$ is varied while $\alpha_2=1$ and vice versa. All results are over 10 trials.}
  \label{fig:parasense}
 \end{figure}

%\vspace{-10pt}
In this section, we study the effect of hyper-parameters on the recognition performance. 
We only report results of sensitivity with respect to $r$, $\alpha_1$ and $\alpha_2$ in Fig.~\ref{fig:parasense}. 
We found our recognition performance to be negligibly sensitive to $q$ and $k'$. 
This suggests that the location of the novel-class prototype estimate only 
depends on the number of subspaces ($r$) rather than its dimensionality ($q+1$). 
Similarly, the Markov-chain-based prediction does not 
depend on the number of nearest neighbors $k'$ used for connecting the graph. 
In Fig.~\ref{fig:parasense}(a), the number of subspaces ($r$) is varied, 
keeping rest of the hyper-parameters the same. 
This is done for both the ImageNet (denoted as (I)) and the CUB-200 dataset (denoted as (C)). 
From the plot, it is seen that the performance increases as the number of subspaces increases 
but then decreases after reaching a peak. 
Initially, more linear subspaces in the neighborhood are required for estimating 
the local structure of the non-linear manifold but becomes irrelevant after more supbspaces.

Next, we studied the effects of $\alpha_1$ and $\alpha_2$ on the recognition performance 
for the 1- and 5-shot settings of the ImageNet and the CUB-200 datasets 
as reported in Figs.~\ref{fig:parasense}(b) and ~\ref{fig:parasense}(c), respectively. 
Accordingly, we obtained an optimal performance when $\alpha_1$ or $\alpha_2$ is between 0 and 1. 
From Eq.~\eqref{eq:weigh}, it suggests that the location of the novel-class prototype 
is within the space bounded by the few-shot sample mean ($\bx_n$), 
the subspace projection ($\bc_p$) and the weighted mean of the nearby class prototypes ($\bc_d$). 
For higher number of shots, 
the maxima seems to move towards the right; that is, closer to $\alpha_1$, $\alpha_2$ values of 1. 
This implies more contribution from the few-shot class mean as visible from Eq.~\eqref{eq:weigh}. 
This is intuitive because as the number of shots increases, 
we expect the few-shot sample mean to converge to the class prototype. 
In fact, for shots of 10 and higher, we obtained the maxima at $\alpha_1=\alpha_2=0.9$ on both datasets. 
For low values of $\alpha_1$, $\alpha_2$ (less contribution of the few-shot sample mean), 
we observed a dip in performance, even sometimes worse than the NA baseline. 
This suggests that the contribution of the few-shot mean 
is important in estimating the novel-class prototype. 
From the plot, we see that the setting $\alpha_1=0$, $\alpha_2=1$ 
produces much better performance as compared to $\alpha_1=1$, $\alpha_2=0$. 
According to Eq.~\eqref{eq:weigh}, it means that the contribution of the projection 
($\bc_p$) is more important compared to contribution of nearby prototypes ($\bc_d$).
%\vspace{-10pt}%%%%
\section{Conclusions}
%\vspace{-10pt}
In this paper, we have introduced a new setting in few-shot learning that assumes access 
to only the base-class prototypes. To address this problem, we used the structural arrangement of the class prototypes on a manifold, firstly to estimate the novel-class prototypes 
and secondly to induce an absorbing Markov-chain for test-time prediction. 
From the experiments, it is evident that our proposed method improved over the no-adaptation baseline but there is still a lot of room for improvement to reach the oracle-level performance. Therefore, our results serve as a benchmark for future researchers to work upon.
\subsubsection*{Acknowledgments.} This work was supported in part by the National
Science Foundation under Grant IIS-1813935.
Any opinion, findings, and conclusions or recommendations expressed in this material are
those of the authors and do not necessarily reflect the views of the National Science Foundation. We also gratefully acknowledge the support of NVIDIA Corporation for
the donation of a TITAN XP GPU used for this research.

%
%
%

\bibliographystyle{splncs04}
\bibliography{main}
\end{document}